\RequirePackage{fix-cm}

\documentclass[smallextended]{svjour3}       
\smartqed  
\usepackage{graphicx}
\usepackage{array}
\usepackage{multirow}
\usepackage[dvipsnames]{xcolor}

\usepackage[sort,authoryear]{natbib}
\usepackage[hidelinks]{hyperref}
\begin{document}

\title{A robot that counts like a child: a developmental model of counting and pointing
	
}


\author{Leszek Pecyna         \and
        Angelo Cangelosi         \and
        Alessandro Di Nuovo 
}


\institute{L. Pecyna \at
              Centre for Robotics and Neural Systems,
              University of Plymouth,
              Plymouth, United Kingdom\\
              \email{leszek.pecyna@plymouth.ac.uk}           
           \and
           A. Cangelosi \at
              University of Manchester, Manchester, United Kingdom and AIST-AIRC, Japan\\
         \and
	         A. Di Nuovo \at
	         Sheffield Robotics,
	         Sheffield Hallam University,
	         Sheffield, United Kingdom\\
}

\date{Received: date / Accepted: date}

\maketitle
\begin{abstract}

In this paper, a novel neuro-robotics model capable of counting real items is introduced. The model allows us to investigate the interaction between embodiment and numerical cognition. This is composed of a deep neural network capable of image processing and sequential tasks performance, and a robotic platform providing the embodiment - the iCub humanoid robot. The network is trained using images from the robot's cameras and proprioceptive signals from its joints. The trained model is able to count a set of items and at the same time points to them. We investigate the influence of pointing on the counting process and compare our results with those from studies with children. Several training approaches are presented in this paper all of them uses pre-training routine allowing the network to gain the ability of pointing and number recitation (from 1 to 10) prior to counting training. The impact of the counted set size and distance to the objects are investigated.
The obtained results on counting performance show similarities with those from human studies.

\end{abstract}

\section*{Introduction}
\label{intro}

The objective of the study is to examine if a neuro-robotics model, using pointing strategies, can replicate the process of 4-year-old children learning to count and, if so, to what extent. As the model uses pointing gestures while counting, we want to investigate their contribution to the counting process and compare the results against those from studies with children.

In order to develop a model capable of counting it is important to state a definition of counting itself. It is not simply the recitation of number words. This is a process of enumeration that allows finding the number of items that the model (or a child) is asked to count.
\cite{gelman1978child} defined five counting principles: the one-to-one principle - each item gets a unique tag; the stable order principle - across trials the tags are ordered in the same sequence; the cardinal principle - the last tag used in a sequence represents the number of
items; the abstraction principle - there are no restrictions about the kind of objects to be
counted; the order irrelevance principle - the items can be counted in any order.

Children familiar with number words and their order are not necessarily capable of counting \citep{wynn1992children, wynn1990children,le2006re}. \cite{wynn1990children,wynn1992children} found that children first memorise a short counting list and it takes around a year till they are able to use counting to determine the cardinality of a set, before that they often violate the counting principles. Around the age of 4, children start to use counting principles and understand the meaning of numbers above the subitizing range \citep{le2006re,le2007one}.


There is long-standing evidence that gestures like pointing and finger counting are an integral part of the development of children's number knowledge \citep{graham1999role,goldin2014gestures,alibali1999function,fischer2012finger,andres2008finger}. It has been confirmed in many studies that pointing, touching or moving objects have a beneficial effect on the counting performance of children \citep{alibali1999function, fuson1982acquisition, gelman1980young, graham1999role, schaeffer1974number}. Preventing children from pointing can disrupt their counting and they end up emitting an indefinite stream of number words or stop counting at all \citep{schaeffer1974number}. According to \cite{alibali1999function}, 4-year-old children's counting performance is facilitated both by pointing and touching gestures made by the child themselves or by someone else. The effect of pointing on counting accuracy has been found particularly strong for children around 4 years old, in contrast to 2- and 6-year-olds \citep{saxe1981gesture}.

There are several proposed explanations of the influence of gestures on the learning to count process. According to one of the earliest proposals, pointing and touching gestures while counting are helping in keeping track of counted objects \citep{schaeffer1974number}. They might work as a memory register identifying the currently counted object and, indirectly, objects counted so far. This may help with the implementation of one-to-one correspondence (assigning exactly one distinct counting word to every counted item) by individuation of items \citep{gelman1980young}. Another hypothesis is that pointing and touching play a coordinative role, synchronising the production of the number words with each counted item so that the one-to-one principle is preserved \citep{fuson1982acquisition}.

Given the importance of numerical cognition in several disciplines, there have been several computational models of the enumeration process. One of the first was developed by \cite{amit1988neural}, where an artificial neural network was counting clock chimes. There were, however, not many attempts to implement a counting model using cognitive Developmental Robotics (DR) principles. The DR seems naturally suitable to study embodied basis of number sense \citep{DiNuovo2015,di2019development} as the robot can interact with the environment, and sensorimotor data can be used in the training process. In several studies \citep{Vivian2014,DiNuovo2014,DiNuovo2014a,DiNuovo2015,di2017embodied,shu20884,pecyna2018Adeep}, DR models were used to explore whether the association of finger representation (finger counting) with number words (and other visual or auditory inputs) could serve to bootstrap numerical knowledge. See \cite{di2019development} for a more detailed review of numerical cognition in embodied artificial systems and the interconnection with developmental psychology and neuroscience.

Among cognitive modellers, deep learning architectures and algorithms are becoming popular as they represent a new efficient approach to building many layers of information processing stages in deep architectures for pattern classification and for feature or representation learning \citep{LeCun2015}. Recent studies, i.e. \cite{dinuovo_ijcnn2020,dinuovo_nature2019} show that sensory-motor information (finger positions) from the child-robot iCub can increase the efficiency of deep learning models in the recognition of spoken and written digits from real world databases. The results of the simulated training show several similarities with studies in developmental psychology, such as a quicker creation of a uniform number line. A discussion on deep learning models for numerical cognition can be found in \cite{testolin_fhn20}.

The first model using DR for object counting was presented by \cite{rucinski2012robotic}, where the neural network used sensorimotor data from an iCub humanoid robot pointing to the counted items. The model was not producing pointing gestures itself (they were used as an input) and visual input was simplified to a one-dimensional vector with 20 units (activation of a particular unit corresponded to the existence of the item in that location). Another model used for item counting was presented in \cite{pecyna2018influence} where the network was able to produce a pointing output by itself (the visual input was still a one-dimensional vector). In their model, a double pre-training process was introduced and its influence, as well as the influence of gesture production, was investigated.

In this paper, we present a new DR model that uses real images from robot cameras together with robot sensorimotor data (joint angles corresponding to pointing) to learn to count a set of table tennis balls. Another novelty of this model is that it is not only producing pointing gestures by itself but also the effect of gestures is visible in the input image in the form of a pointing hand (added to the captured or partially generated image of table tennis balls). This makes pointing more realistic and similar to children one, as they can see their hand when they point. To process the image of counted objects the model uses a deep neural network. The model is counting the objects and providing the information about the number of presented items. We compare the results of our model with those obtained with children \cite[particularly those presented in][]{alibali1999function} in various training conditions.
One of the objectives of our model is to check how pointing gestures affect a counting process and if this is in line with what was observed in children. In this paper, by the term gestures we mean hand movements/positions in general, we use expressions like gesture output to name a hand position output in our model. As the hand movements that we are replicating with a robot are indicating (pointing to) objects, those movements correspond in our studies to pointing gestures. Similarly in \cite{alibali1999function}, the term gestures was used for pointing to or touching objects to indicate them.

The paper is organised as follows. First, we present the robotic model and its architecture. In the further section, we explain how the training data was collected and pre-processed. This is followed by the explanation of the training routine. Next, we present several simulation studies where we train the model in different ways (aimed to reassemble children counting ability strategies). The most convincing training approach is analysed in comparison with results from developmental psychology studies to verify its fitting with the children data. We consider aspects such as the influence of pointing, distance to objects or size of the counted set. The article finishes with the summary and conclusions.

\section{Model Description}\label{descr}
The main task of our model is to generate number words (one word at one time step) and finish this recitation (counting) when the model reaches the number of objects presented in the image. The network can use generated by the robot pointing hand movement data (sensorimotor and images) while performing this task.

The general training idea and the scheme of the network is similar, on some level, to the one presented by \cite{pecyna2018influence}. One of the differences is that in this work the network is trained with real images collected using a robot camera (and a deep neural network architecture). The input pictures present a set of table tennis balls located on a black surface. In these images, the robot hand pointing to those objects (to one place at each time step) can be also included\footnote{We also tested the model without that feature (when gestures are produced but the hand is not visible). That case is similar to what was presented by \cite{pecyna2018influence}. We found better performance with hand visible but these analyses are not in the scope of this paper.}. The hand position can be based on the gesture output generated by the network or can be added to the visual input in the pre-processing (``puppet pointing'' condition, described in Section~\ref{training}). Pointing is restricted to 11 positions corresponding to 11 possible columns in which objects can be positioned. More about how data was collected can be found in Section \ref{data}. The network is trained using backpropagation algorithm and backpropagation through time for the network recurrences.

\begin{figure}[tb]
	\centering
	\includegraphics[width = 0.63\textwidth]{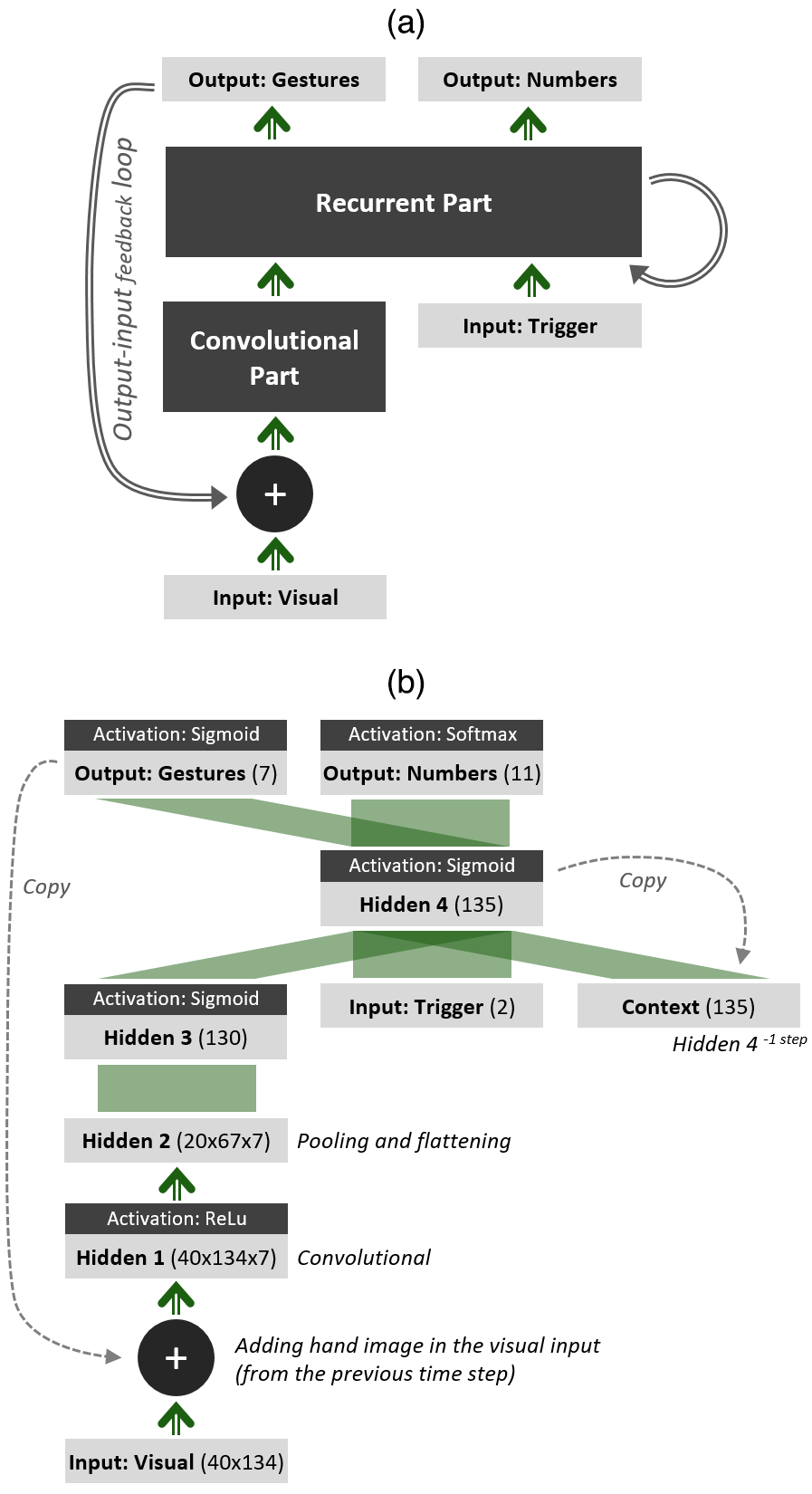}
	\caption{Model architecture. (a) General scheme of the model. It presents all the inputs and outputs of the network and defines parts of network responsible for particular tasks: Convolutional Part - image processing, Recurrent Part - generating sequential processes. (b) Network architecture - polygons represent all-to-all connections between layers of neurons; other connections between layers are indicated by arrows. Values in the brackets represent the sizes of the layers. A sign: ``+''  is a function combining input image with a hand image based on the gesture output (see more in the text).}
	\label{fig:model_general}
\end{figure}

As the model uses real images, convolutional network layers were used in our architecture. The general structure of the model is presented in Fig.~\ref{fig:model_general}a. The network is composed of two main modules: the Recurrent Part, which is an Elman recurrent neural network responsible for sequential processes and the Convolutional Part where the image is processed. 

An Elman recurrent neural network is a Simple Recurrent Network (SRN) where the hidden layer activation values are copied (during each time step) into the context layer. In the following time step, this context layer is used as an input to the hidden layer. In this way, even if all the other inputs stay the same, the network can produce a different output (based on its previous activation status) and is able to learn sequential tasks \citep{elman1990finding}.
There are more complex and powerful recurrent neural network currently used, such as Long Short-Term Memory (LSTM) networks. These networks are designed to learn long-term dependencies in the input sequence \citep{hochreiter1997long}. Learning these long-term dependencies is difficult to obtain with classical SRNs. However, in the case of our problem, we used only up to 15 times steps sequences, and the following step in the sequence can be determined (in the case of counting words) directly from the previous one. In this scenario, the use  of a  more complex structure is unnecessary to obtain the counting and pointing behaviour. Elman-type networks were successfully used for counting purpose before by \cite{rodriguez1999recurrent} to count letters, by \cite{rucinski2012robotic,pecyna2018influence} to count elements in the simplified visual input and by \cite{Vivian2014} to count number words.

Convolutional neural networks are designed to process image data (or in general multiple arrays) and recognize patterns in it. The input of the convolutional layer is divided into local patches which are connected to the next layer (feature map) via a set of weights (trainable filters) \citep{LeCun2015,scherer2010evaluation}. A deep architecture with convolutional layers represents the current state of the art in computer vision, and is inspired by the biological organisation in the visual cortex in animals and humans \citep{dinuovo_nature2019}. A typical convolutional network uses a pooling layer after the convolutional one. A pooling layer serves to reduce the spatial size of the representation and the amount of computation in the network \citep{dinuovo_nature2019}, and also allows the network to achieve spatial invariance \citep{scherer2010evaluation}. A max pooling layer (the most common type of pooling layers, used in our studies) computes the maximum of a local subregion of units in the feature maps. 

Apart from the loop inside the Recurrent Part, there is another loop in the network: an output-input feedback loop for gestures. This loop allows the network to see the hand/fingers in the image at the position generated by the network in the previous time step. The gestures' addition function converts the pointing angular position from the previous time step into a hand image and adds that image to the visual input with counted balls. In this way, our model can generate the pointing by itself.

	\subsection{Network architecture}\label{model_structure}
	A detailed architecture of the neural network, which is the implementation of our general scheme, can be seen in Fig.~\ref{fig:model_general}b. All layers and their sizes are presented. The hyperparameters are the size of the hidden layers, kernel size, number of filters, type of activation function and values of learning rates (see Table~\ref{tab:parameters}). They have been adjusted using hyperparameter search.
	To simplify that process, we first adjusted the elements of the network responsible for pointing (gesture production and image processing) as in gesture pre-training (see Subsection \ref{pre}). Then, we adjusted other parameters running the simulations with the full network. This way we decreased the number of dimensions scanned at one time in the hyperparameter's space.
	
	\begin{table}[tb]
		\caption{Chosen training parameters}
		\centering
		\begin{tabular}
		{>{\centering}m{0.52\columnwidth}>{\centering\arraybackslash}m{0.11\columnwidth}}
			\noalign{\smallskip}\hline\noalign{\smallskip}
			\textbf{Parameter} & \textbf{Value} \\
			\noalign{\smallskip}\hline\noalign{\smallskip}
			Amount of convolutional filters& 7 \\
			Size of a kernel& 3 \\
			\noalign{\smallskip}
			Size of Hidden 3 layer& 130 \\
			\noalign{\smallskip}
			Size of Hidden 4 layer& 135 \\
			\noalign{\smallskip}
			Gesture pre-training learning rate& 0.004 \\
			\noalign{\smallskip}
			Number pre-training number learning rate& 0.002 \\
			\noalign{\smallskip}
			Number pre-training gesture learning rate& 0.001 \\
			\noalign{\smallskip}
			Main training gesture learning rate& 0.002 \\
			\noalign{\smallskip}
			Main training number learning rate& 0.001 \\ 
			\noalign{\smallskip}\hline
		\end{tabular}\label{tab:parameters}
	\end{table}

	The Convolutional, Pooling and Hidden 3 layers visible in Fig.~\ref{fig:model_general}b compose the Convolutional Part of the network from our general model scheme (see Fig.~\ref{fig:model_general}a) responsible for image processing. A Hidden 4 and its Context layer constitute the Recurrent Part (from the general scheme) allowing number recitation and sequential gesticulation.
	\subsection{Inputs and outputs}\label{input}
	As visible in Fig.~\ref{fig:model_general}, the network has two sets of inputs and two sets of outputs described as follows:
	\begin{itemize}
		\item Trigger input: A vector with two units. Its role is to indicate when the counting and/or pointing process should be performed. The network is suppose to produce zeros whenever the trigger unit corresponding to counting is off, and should keep the hand in the ``base'' position whenever the second trigger unit (corresponding to pointing) is off (it learns these behaviours in the training process). The desired counting is produced when both trigger units are on. The trigger value remains constant throughout the whole time sequence in the training and testing data sets. The trigger vector together with additional information in the image tells the network what task is supposed to perform. All possible simulated skills are described in Section \ref{training} and listed in Table~\ref{tab:modalities}.
		\item Visual input: A grey-scale image of size 40 by 134 pixels. In the image, a number of table tennis balls (from 0 to 10) are presented in different configurations (more about how image data was collected and possible positions of balls can be found in Section \ref{data}). Aside from the picture of balls, additional trigger information is included in that input. This consists of a single-pixel wide horizontal white line at the right of the image. It informs the network if the image will be used for pointing or counting, acting as a light bulb to indicate if counting should, or should not be performed\footnote{Having this information in the visual input is useful for the model because the Convolutional Part (image processing part) can learn to react differently whenever the image has to be processed or not.  We did not include other trigger data in the visual input because of the pre-training method we initially (and as a subsidiary study) used, where only part of the network was trained and it was not necessarily containing the visual processing units.}.
		\item Number output: A vector with 11 units. One-hot coding is used for simplifications. Particular units correspond to numbers from 0 to 10.
		\item Gesture output: A proprioceptive signal from the iCub robot that the network is trained to reproduce. It consists of 7 units corresponding to angles of the first 7 arm joints. The output represents the hand position.
	\end{itemize}
	\section{Training dataset}\label{data}
	To simplify the task we used white objects on a black background so that colour pictures do not have to be processed. We chose table tennis balls as the objects to be counted by the robot in the experiment.
	
	The robotic platform we used was an iCub humanoid robot. The iCub is a child-like robot designed to support cognitive developmental robotics research \citep[for more details about the robot see][]{metta2010icub}.
	
\begin{figure}[tb]
		\centering
		\includegraphics[width = 1\textwidth]{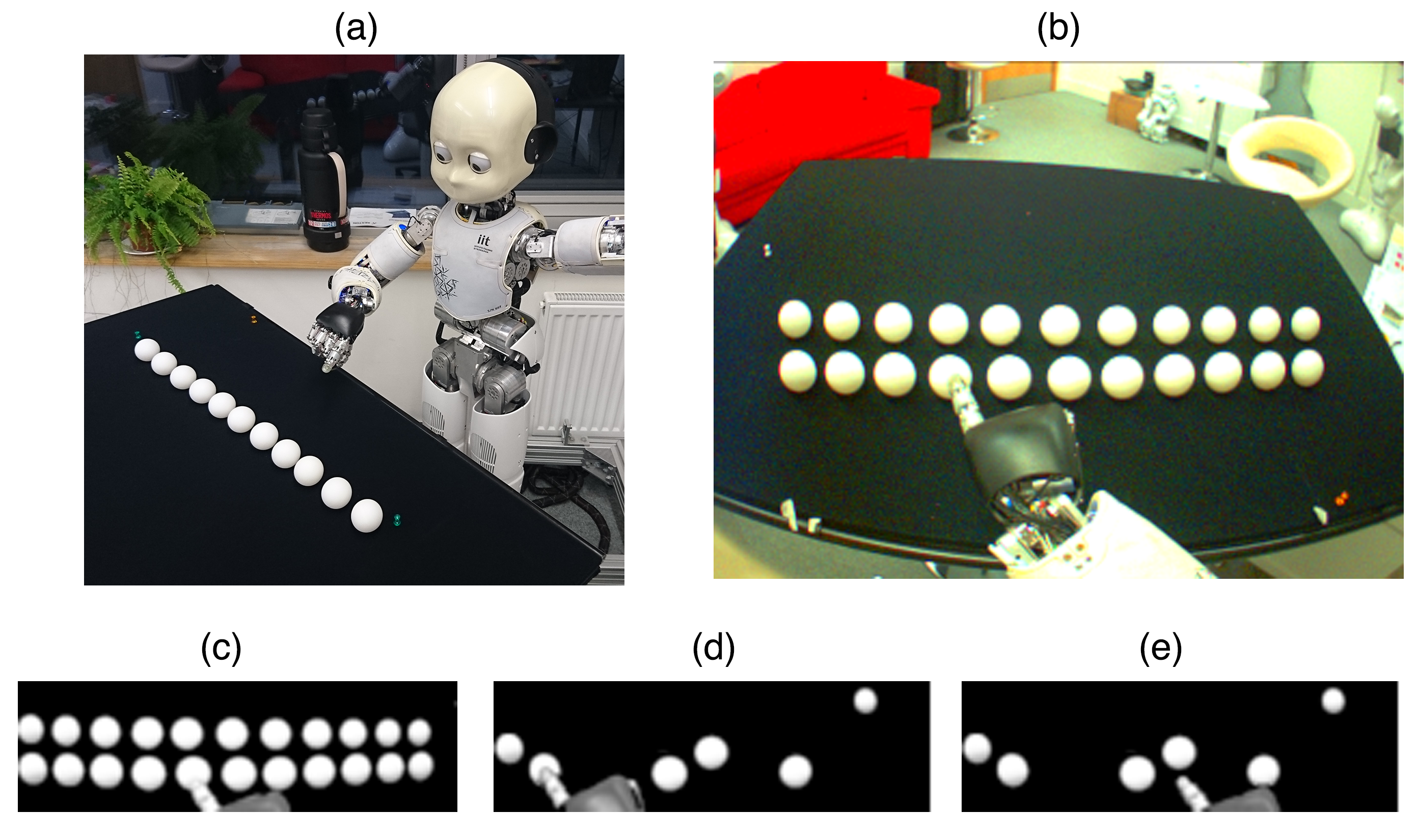}
		\caption{Experimental set up and input images. (a) iCub humanoid robot pointing to the table tennis balls - robot configuration. (b) iCub camera view of robot hand pointing to the objects. (c) Processed image: cut image (row 1 and 3) with a pointing hand. (d) Generated image - pointing to the second position. (e) Generated image - pointing to the sixth position.}
		\label{fig:experiment_pointing}
	\end{figure}

	The arrangement of the experiment and the robot position during the data collection can be seen in Fig.~\ref{fig:experiment_pointing}a. Because of the limitations of the robot joints' ranges, the maximum number of objects in one horizontal row was chosen to be 11. There were 5 possible rows in which balls could be located (altogether 55 possible positions). For simplification, these 11 possible left-right positions were located in such places, for each row, that in the robot cameras they were one over another (i.e. the first position of the lowest row was under the first position of the second and further rows). In this way, there were exactly 11 distinct possible horizontal positions to which robot could point.
	
	The iCub robot was pointing to the lowest row of balls. The image of balls and the iCub hand were collected together with sensorimotor data (joint angles). In Fig.~\ref{fig:experiment_pointing}b you can see the robot camera image when the iCub was pointing to the fourth position and all 11 balls were located in both the first and the third row.
	A similar image of the robot hand was collected when there was no ball on the stand so that the robot hand can be easily isolated from the picture for each of the 11 positions (hand orientation was different in each position).
	
	The pictures were cropped and their resolution was decreased to 134 by 40 pixels. The example of such cropped area can be seen in Fig.~\ref{fig:experiment_pointing}c, where the iCub points to the fifth position. Finally, a data augmentation function to generate new images of balls was made. This function allows us to generate images with 0 to 10 balls in the random positions (but never in the same vertical position - so they do not cover one another and pointing is not repeated in one place). Another function allows the addition of a hand image to the picture pointing to any of possible 11 locations. Examples of these generated pictures can be seen in Fig.~\ref{fig:experiment_pointing}d and e.

	The sensorimotor data of the robot's joints was collected to train the network to produce the pointing gestures. In order to collect the joints' data together with images from the cameras, we needed to block the iCub torso angles (and head angles) to avoid having different pictures for different positions. The highest range for the arm (range for pointing left to right) was when the torso jaw angle was set to -14$^\circ$ (this caused the right shoulder joint being closer to the robot centre). The head was rotated 14$^\circ$ so that the robot was looking forward; eyes were set down (25.182$^\circ$) to make the robot look at the objects (iCub posture is visible in Fig.~\ref{fig:experiment_pointing}a).
	During pointing the arm angles were changing, we used 7 first angles of the arm joints for the training (following 9 angles are responsible for fingers positions which were fixed during the experiment).
	
	In the case of the simulated skills where gestures are not required, we defined the base position which is a set of angles, that the network should produce, representing the position of the arm down beside the iCub body.
	
	It should be noted that the data collection process was not implemented in real time. This was because the training process would have required a long time, as we would have to organise table tennis balls in many configurations to train the model. However, after the network is trained it could be used directly with the iCub robot after providing a proper interface.
	Such interface would have to meet many criteria (like proper cropping of the image, and taking care of a hand movement - as only final positions are given by the network) but its implementation seems fully feasible without changing the model architecture. 

\section{Simulated skills}\label{training}
People can use different, previously acquired, skills to learn and perform more complex tasks. In the case of counting, psychological studies show that children can recite short number list (acquire a list of tags) prior to the acquisition of counting skills \citep{wynn1992children,le2006re}. It takes around a year before children start to use known number words for counting at the age of around 3.5 years \citep{wynn1990children,wynn1992children}. Even before that, they already possess complex motor skills e.g. children successfully perform reaching task before their age of 28 months \citep{bertenthal1998eye}. As shown by \cite{alibali1999function} and \cite{graham1999role} when children count and are not instructed to point or not most of them perform pointing to the objects or even touch them. Thus, to replicate the children learning to count process, the presented model should be able to perform and learn these two simpler skills (pointing and recitation). These skills should also be acquired in an appropriate (resembling children training) order.

In \cite{alibali1999function} children were asked to perform a few different counting tasks.
In their studies, 20 children (10 boys and 10 girls) in the age range from 4 years, 3 months to 5 years, 4 months (mean = 4 years, 8 months) were asked to count sets of plastic chips on strips of cardboard (chips were 1 inch in diameter and were pasted approximately 1/2 inch apart). The sets ranged in size from 7 to 17 chips. The authors conducted the experiments in several conditions: (1) no instructions - there was no instruction given regarding pointing or touching items while counting; (2) child point - children were instructed to point (but not touch) each counted cheap; (3) child touch - children were instructed to touch each chip as they counted it; no gesture - children were told not to point to or touch the chips; (4) puppet point - the puppet (pink pig puppet, controlled by the experimenter) pointed to the chips as the child said the number words; (5) puppet touch - the puppet touched each chip as the child said the number words; (6) puppet incorrect -  puppet made errors in pointing.
The results from those studies were analysed from the point of view of the influence of pointing and touching on counting.

In our studies, we aim to replicate some of those conditions, mainly: child point, no gestures and puppet point (touch conditions are partially considered in Subsection \ref{dist}).
To do so, we defined (together with basic aforementioned skills and the quiescence case - ``Do nothing'') 6 particular simulated skills that our model can be trained to perform:
\begin{enumerate}
	\item Do nothing: the model is supposed to produce 0 throughout all of the time steps and keep the gesture output corresponding to the base position.
	\item Pointing: the model is supposed to produce 0 throughout all of the time steps and point to objects from left to right.
	\item Recitation: the network is trained to produce words from 1 to 10 through the first 10 time steps and keep the gesture output corresponding to the base position.
	\item Counting with pointing: the model is producing number words corresponding to presented objects (it counts the objects), it should finish counting when reaches the last object. At the same time, it is pointing to these objects.
	\item Counting without pointing: the network is producing number words as in the above task but it is trained to keep gesture output at the base position.
	\item Puppet (puppet pointing): this is the same task as the one above (network is counting and a hand is supposed to stay in the base position), the only difference is that the input image contains a hand that is pointing to the objects from left to write (each object at the time step). The image of the hand is the same as in the ``Counting with pointing'' case but it is not produced by the network and always provided in the perfect manner.
\end{enumerate}
The usage of particular simulated skills is determined by the input triggers as shown in Table~\ref{tab:modalities}.

\begin{table}[tb]
	\centering
	\caption{Simulated skills}
	\label{tab:modalities}       
	\begin{tabular}
	{>{\centering}m{0.022\columnwidth} >{\centering}m{0.12\columnwidth} >{\centering}m{0.12\columnwidth} >{\centering}m{0.12\columnwidth} >{\centering\arraybackslash}m{0.18\columnwidth}}
		\hline\noalign{\smallskip}
		\textbf{No.} & \textbf{Visual trigger} & \textbf{Gestures trigger} & \textbf{Number trigger} & \textbf{Task} \\
		\noalign{\smallskip}\hline\noalign{\smallskip}
		1 & False & False & False & Do nothing\\
		\noalign{\smallskip}\noalign{\smallskip}
		2 & True & True & False & Pointing\\
		\noalign{\smallskip}\noalign{\smallskip}
		3 & False & False & True & Recitation\\
		\noalign{\smallskip}
		4 & True & True & True & Counting with pointing\\
		\noalign{\smallskip}
		5 & True & False & True & Counting w/o pointing\\
		\noalign{\smallskip}
		6$^a$  & True & False & True & Puppet\\
		\noalign{\smallskip}\hline\noalign{\smallskip}
		\multicolumn{5}{l}{\footnotesize{$^a$ The same as 5 but different image input (hand visible).}}
	\end{tabular}
\end{table}

In all of the counting skills (and the recitation one), the network, after finishing counting (or recitation), should keep the last value of the number word throughout the remaining time steps. We always use  15 time steps in our training and testing procedures.

The above skills are not trained separately, e.g. in the final training (after the modal was pre-trained) we train the network to keep the ability to point and to recite but we add counting to the training. A training batch, in that case, is composed of 1, 2, 3, 4 sub-batches corresponding to the aforementioned simulated skills.
It is important to mention that, even if our model was never trained with some of the skills, it can be tested on how it performs them at any stage of the training.

\section{Experimental studies with different simulated skills}\label{human}
In this section, we will present training strategies and their results where we set out to train our network, using methods that we assume are similar to a child's counting development. The network undergo pre-training before the main training (as described below). Upon completion, the network is able to point and recite numbers presented to it. We made some training assumptions and we chose three training strategies (Study 1, 2 and 3) presented in the following subsections.
	
	\subsection{Pre-training sessions}\label{pre}

\begin{figure}[tb]
		\centering
		\includegraphics[width = 0.63\textwidth]{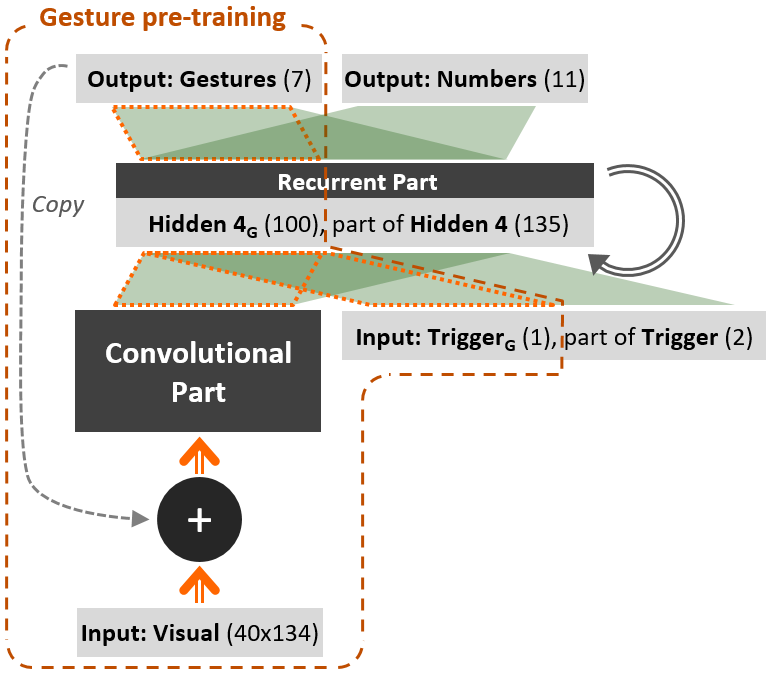}
		\caption{Partially simplified scheme of the network architecture. It is indicating which part of the network is trained in the gesture pre-training. Polygons represents all-to-all connections between layers of neurons: dotted ones - trained connections in the gesture pre-training, filled ones - all connections trained in the final training (and the recitation pre-training).}
		\label{fig:pre-trainings}
	\end{figure}

	In our studies, we introduce two preliminary training sessions that might help in the final training. One is pointing pre-training and another is number recitation pre-training. To keep similarity with the child learning process, we first train the network to point to the objects, and later to recite the numbers. These skills are also acquired in this order by children \citep{bertenthal1998eye, fuson1988children, wynn1992children}. It was already found that pre-training sessions like these (for pointing and number recitation) can boost the network performance and the training speed \citep{pecyna2018influence}\footnote{Also a 
	few subsidiary tests were performed using the presented model and they showed that the gesture pre-training increases the final training speed significantly (where the recitation pre-training had positive influence only if used together with the gesture pre-training).}. In more detail, these preliminary training sessions are described below:
	\begin{itemize}
		\item Gesture pre-training: As we want to keep particular regions of the network responsible for this ability, we specify a part of the network that is trained in this pre-training (Fig.~\ref{fig:pre-trainings}). This part is trained to perform pointing from left to right over the time steps (it does not have a number output) and in the case of a zero gesture trigger, it is supposed to do nothing. This pre-training is similar as the one introduced in \citep{pecyna2018influence}.
		\item Recitation pre-training: In this training, we use the whole network\footnote{At the beginning of our research we were training the model to recite in a similar manner as in the case of gesture pre-training - only part of the network was trained. We found, however, that such a pre-trained network when trained to count, immediately loses the ability to recite (even if it is included in the final training set of simulated skills). This is likely because the model was never trained to produce any type of gesture output (in the case of recitation in the final training and current recitation pre-training the network is producing gestures - base position) and through backpropagation, gesture output error modifies the weights responsible for number recitation.}. As we want the network to keep the pointing ability, we train it to perform 3 skills (defined in Section \ref{training}): 1, 2 and 3 (Do nothing, Pointing and Recitation).
	\end{itemize}
	
	\subsection{Training description}
	
	When we generate our training data we create 1000 images for each numerosity i.e 1000 images with one object and 1000 images with two and other number of objects (randomly distributed in the picture). We define a batch as a set of images (and other corresponding training inputs and outputs) composed of 10 elements covering all numerosities (from 1 to 10). Thus, our training data is composed of 1000 batches. Each batch also covers all simulated skills the network is trained to perform during the particular training session. An epoch is defined as a training session throughout all the batches one time (an iteration - training thorough one batch).
	
	In each training session, we also produce a test set, which is, as well, randomly generated (random positions of balls), it is composed of 50 batches. We test our network with such a set after every 50 iterations of the training (this allows us to check if the model is not biased to respond correctly only for the data it was trained with). A test set is always composed of all 6 simulated skills (60 elements in one test batch i.e. 6 skills with 10 different numerosities) so that we test the model performance for all of them.
	
\begin{figure}[tb]
		\centering
		\includegraphics[clip, trim=4.5cm 9cm 5cm 9cm,width=0.63\textwidth]{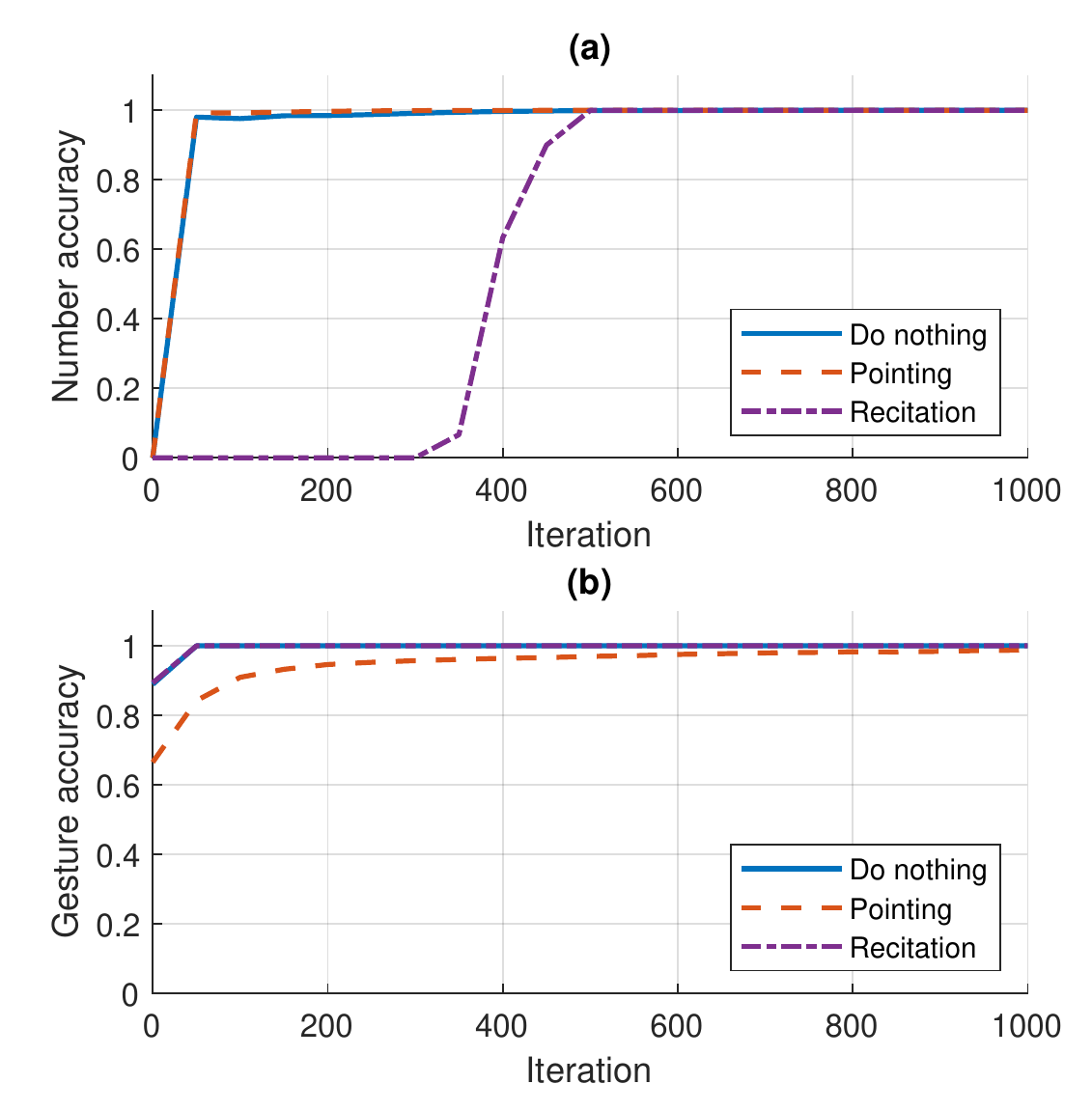}
		\caption{Accuracy development, average from 30 simulations for the second stage pre-training (recitation). The model was trained to recite the numbers and to keep previously acquired abilities (Pointing and Do noting). Tested for first 3 simulated skills, as described in Section \ref{training} (test set used every 50 iterations). (a) Accuracy of a number output. (b) Accuracy of a gesture output.}
		\label{fig:1234_pre_num}       
	\end{figure}

	The training simulations presented in this section are always preceded by both pre-training procedures described in Subsection \ref{pre}. First, the network is trained for 2 epochs (2000 iterations) to produce pointing and then for the following 1 epoch, a number recitation training is conducted.
	After such double pre-training, our modal is performing almost perfectly in these two skills (and Do nothing one). Please see the number recitation pre-training in Fig.~\ref{fig:1234_pre_num}.
		
	After the pre-training sessions, we conduct the final training where the network is trained to count. Here we use 3 approaches. First, we assume that the network is trained to count and to point to the objects (but is not trained to perform counting without pointing nor to count with a puppet). Thus, in the final training the network is trained with 4 simulated skills (keeping the ability of performing the first 3 ones): 1, 2, 3 and 4 (Do nothing, Pointing, Recitation and Counting with pointing). A more detailed description of this training and the results are presented in the next subsection.
	
	Secondly, we train our model with all possible simulated skills (1 to 6 as presented in Section \ref{training}). Thus, the network is simultaneously trained to count with and without pointing and to count when the puppet is performing the gestures. This is presented in Subsection \ref{count_all}.
	
	Finally, we try to adjust, at some level, the number of particular simulated skills in the training set (how much counting with pointing and how much without it should appear in the training etc.) to resemble children counting training process. How that was done and the results of this training are presented in Subsection \ref{count_complex}.
	
	\subsection{Study 1 - Training with pointing}\label{count_w_gest}
	
\begin{figure*}[tb]
		\centering
		\includegraphics[clip, trim=0.5cm 6.8cm 0.5cm 6.5cm,width=1\textwidth]{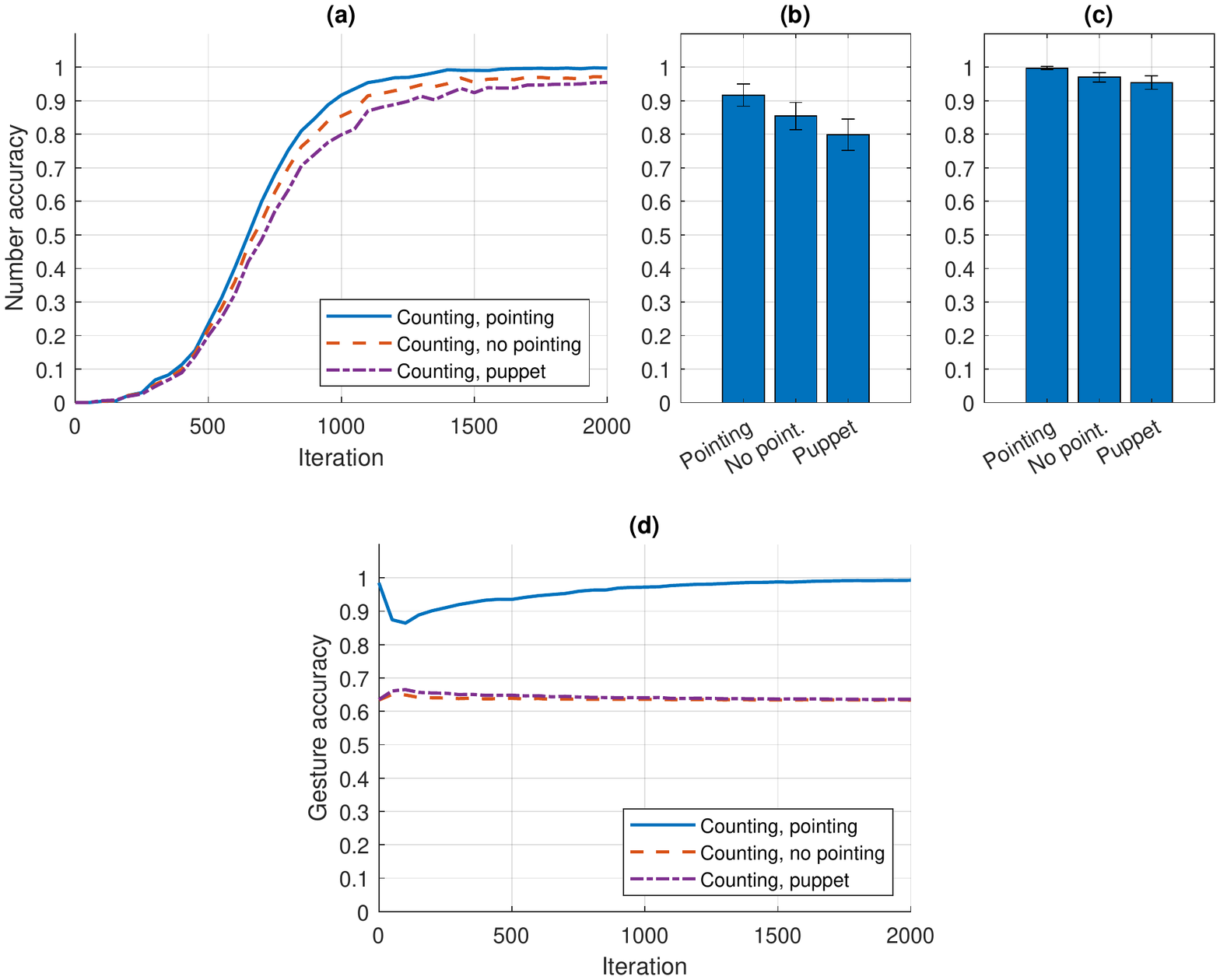}
		\caption{Accuracy of counting (number output) and pointing (gesture output), average from 30 simulations. The model was trained to count with pointing. Tested in 3 conditions: when pointing required, when pointing forbidden and when counting with a pointing puppet. The error bars (in the bar plots) indicate 95\% confidence intervals. (a) Accuracy of counting development (test set used every 50 iterations). (b) Accuracy after 1000 iterations (1 epoch). (c) Accuracy after 2000 iterations (2 epochs). (d) Accuracy of pointing development.}
		\label{fig:1234_num}       
	\end{figure*}

	In this study, we trained the model to perform skills 1, 2, 3 and 4  (Do nothing, Pointing, Recitation and Counting with pointing). The training took 2 epochs (2000 iterations). Batches were composed of 40 elements, one for each possible numerosity (from 1 to 10) for all 4 simulated skills. As described before, the network was tested with a test set of 50 batches and the results are presented in Fig.~\ref{fig:1234_num} for number and gesture output performance. The training sessions were repeated 30 times and the values on the plots represent the average value from these simulations.
    
	We omitted here the results of the skills 1, 2 and 3 (Do nothing, Pointing, Recitation) as these are simple tasks trained in the pre-training sessions. In the final training the performance of the network in these simulated skills was close to 100\%.
	
	As it can be seen, the model is producing the number words most accurately for counting with pointing. 
	This is visible on all of the training stages. This difference is less significant at the end of the training (see Fig.~\ref{fig:1234_num}c). The model was never trained to perform counting without pointing and counting with a pointing puppet, yet its performance in those tasks is still very good. What might be unexpected, is the worst performance obtained in a puppet pointing condition, although this simulated skill seems similar to the one with pointing (visual input contains the hand).
	This might be because in a puppet pointing condition the network gets a perfect visual input and produces some gesture output that might mislead the network (it is used in a recurrent part). This gesture output does not have to be correct as the network might not know how to respond to this specific combination of triggers - it was never trained with it (it was trained to do nothing and to perform counting with pointing and these skills have some of the trigger values similar to this tested puppet skill).
	
	After running a one-way ANOVA (analysis of variance) for the end results' final weights (after 2000 iterations), we can see that there is a statistical difference between our sets (counting with pointing, without pointing and with a puppet) with $F(2,87)=8.57$, $p = 0.004$. A Post Hoc pairwise analysis (Tukey’s test) showed a statistical difference between counting with pointing and without it ($p = 0.032$). It also showed a strong statistical difference between counting with pointing and with a puppet ($p < 0.001$). There was no statistical difference found for the condition without pointing and with a puppet ($p = 0.282$). However, we can see that training without pointing is faster than the one with a puppet i.e. average accuracy reaches higher values faster in the training process (see Fig.~\ref{fig:1234_num}a).
	
	As can be seen in Fig.~\ref{fig:1234_num}d, the only skill in which the network has high pointing performance is counting with pointing. This is easy to explain. In the case of two other skills the model is asked to stay at the base position and at the same time produce number words (it was never trained to do that), thus, having a positive number trigger and the one in the visual input, the network performs, to some level, pointing and that is considered as a wrong output.
	
	These results are partially similar to those obtained on children in \cite{alibali1999function}, where the performance in the no-gesture condition was substantially lower than performance in gesture one (as in the model results). However, in their study, the performance in child pointing condition was only slightly (did not differ statistically) better than the one in a puppet pointing condition. This was different in the case of presented Study 1 where the puppet condition was giving the worse results.

    \subsection{Study 2 - Training with all simulated skills simultaneously}\label{count_all}
    
\begin{figure*}[tb]
		\centering
		\includegraphics[clip, trim=0.5cm 6.8cm 0.5cm 6.5cm,width=1\textwidth]{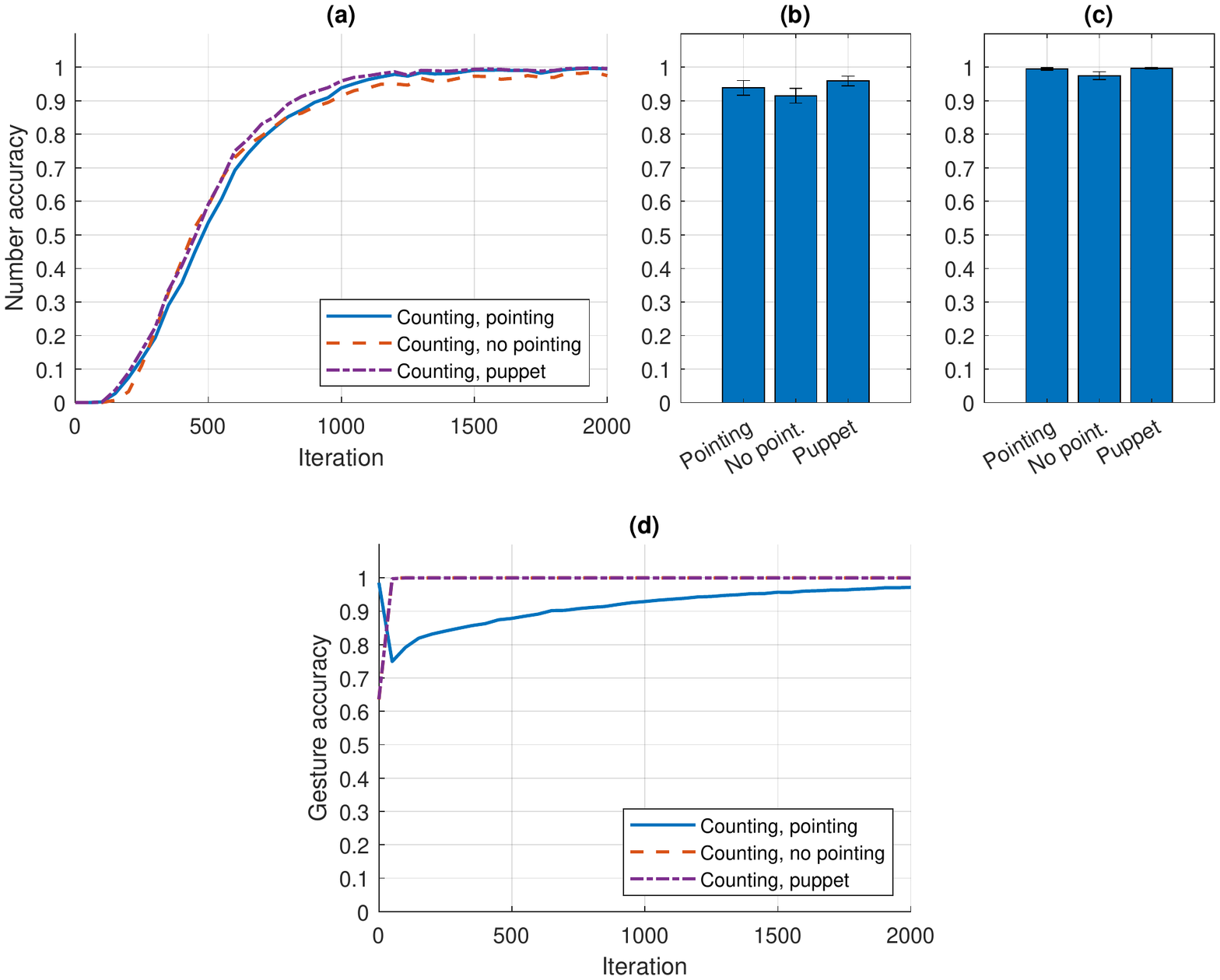}
		\caption{Accuracy of counting (number output) and pointing (gesture output), average from 30 simulations. The model was trained to count with pointing. Tested in 3 conditions: when pointing required, when pointing forbidden and when counting with a pointing puppet. The error bars (in the bar plots) indicate 95\% confidence intervals. (a) Accuracy of counting development (test set used every 50 iterations). (b) Accuracy after 1000 iterations (1 epoch). (c) Accuracy after 2000 iterations (2 epochs). (d) Accuracy of pointing development.}
		\label{fig:123455p_num}       
	\end{figure*}

	In this study, the final training session was composed of all simulated skills 1, 2, 3, 4, 5 and 6. The training took 2 epochs. Batches were composed of 60 elements, one for each possible numerosity (from 1 to 10) for all 6 skills. The network was tested with a test set of 50 batches and the results are presented in Fig.~\ref{fig:123455p_num} for number and gesture output performance. The training sessions were repeated 30 times and the values on the plots represent the average value from these simulations.
	As in Study 1, we omitted skills 1, 2 and 3 (Do nothing, Pointing, Recitation) in our plots and analysis (their performance was close to 100\%).
	
	The difference in performance between the skills is not very high in this study. 
	The performance in a counting with a puppet condition is the best especially in the earlier part of the training (see Fig.~\ref{fig:123455p_num}b for results after 1000 iterations). After a longer training, the performance of counting with pointing reaches similar value and only counting without pointing performance stay lower.
	
	After conducting an ANOVA analysis (for the end results) we found that there is a strong statistical difference between our conditions, $F(2,87)=10.82$, $p < 0.001$. A Post Hoc test showed a strong significant difference between condition with pointing and without it and between condition without pointing and with a puppet ($p < 0.001$). There was no statistical difference found between the puppet condition and the pointing one ($p = 0.964$). 
	Those final results are in some way similar to those with children \citep{alibali1999function}. However, in the case of study with children counting with pointing was slightly (statistically not significantly) over performing the puppet condition. The model results in Study 2 showed the opposite relation between those cases.
	
	As for the pointing performance, we can see in Fig.~\ref{fig:123455p_num}d that the network immediately learns to stay in the base position (in the case of 5 and 6 skill) which is an expected result as this is a very simple task.
	
	\subsection{Study 3 - Dynamic combination of simulated skills}\label{count_complex}
	In this subsection, we will describe training which is a composition of the previously presented conditions. We can assume that a child, to some extent, learns to perform all of the counting tasks that
	our network is trained to do: skills 4, 5 and 6 (Counting with pointing, Counting without pointing and Puppet pointing). We do not know how often these specific tasks are occurring in the case of children.
	
	We made some very general assumptions about such training and tried to estimate the probability of each task during the development of numerical skills. We assumed that at the beginning it is more common that children will only observe a teacher or a parent performing counting, thus, we set the probability of that to 60\%. We also assumed that at this point children will not perform counting without pointing as this is more difficult task. In the final stage of the training (at age 4, as we try to compare our results with children of that age), we made an assumption that children will count by themselves much more often (in 90\% of cases) of which we chose 10\% to correspond to counting without pointing.
	We know that at the age of 4 in around 90\% of cases children will choose to point (or touch the objects) when asked to count \citep{alibali1999function}. This is not the same situation as one child pointing in 90\% of its training trails but we assumed this number might be close to that (thus, assumed 90\% chance to count with pointing at the end of the training).
	These assumptions are very rough as there are no studies which would document this process in such details.

\begin{figure}[tb]
		\centering
		\includegraphics[clip, trim=4.5cm 12cm 5cm 12.5cm,width=0.63\textwidth]{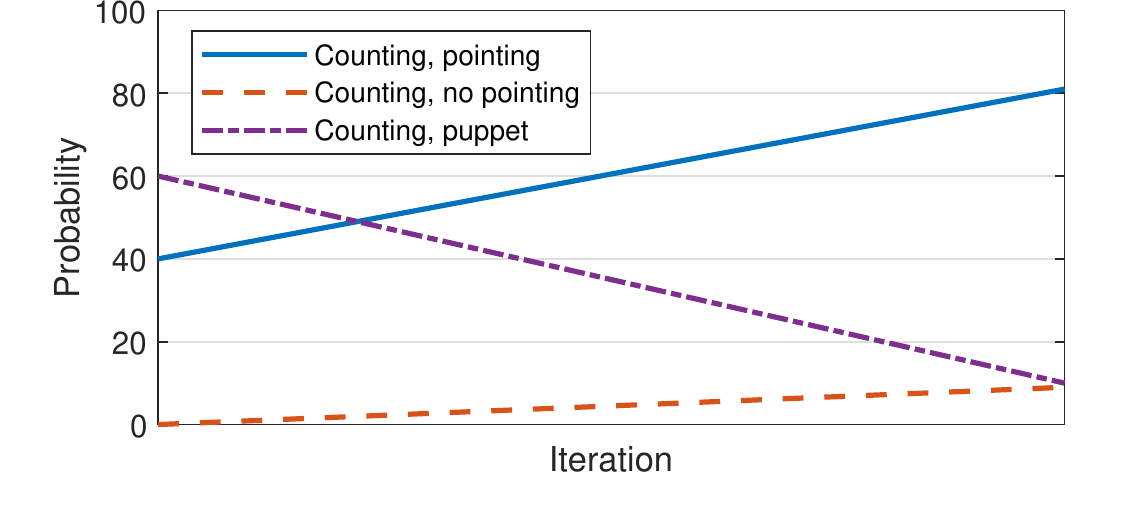}
		\caption{Assumed probability of an appearance of a particular simulated skill: Counting with pointing, Counting without pointing and Puppet pointing in a batch in each training iteration.}
		\label{fig:probability}       
	\end{figure}

\begin{figure*}[tb]
		\centering
		\includegraphics[clip, trim=0.5cm 10.5cm 0.5cm 11cm,width=1\textwidth]{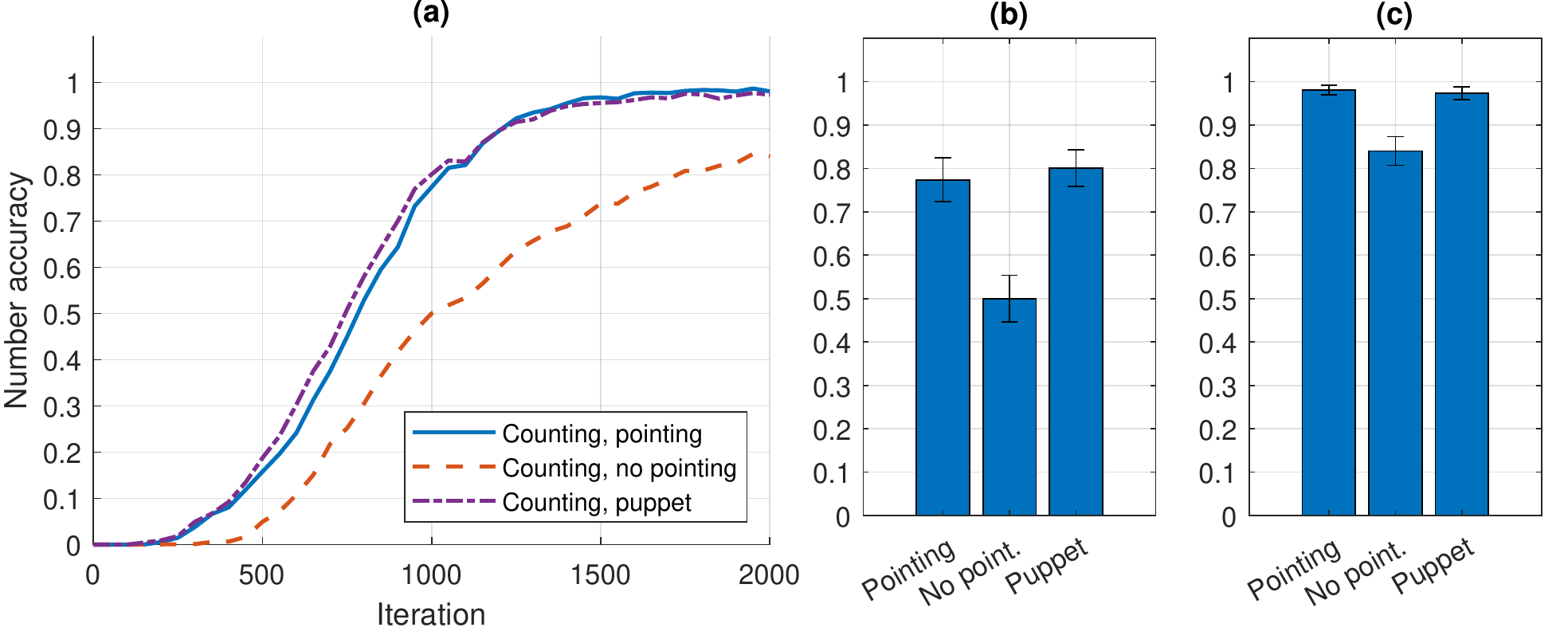}
		\caption{Accuracy of counting (number output accuracy), average from 30 simulations. The model was trained to count in described in this section manner assuming training length of 2000 iterations. Tested in 3 conditions: when pointing required, when pointing forbidden and when counting with a pointing puppet. The error bars (in the bar plots) indicate 95\% confidence intervals. (a) Development curves (test set used every 50 iterations). (b) Accuracy after 1000 iterations (1 epoch). (c) Accuracy after 2000 iterations (2 epochs).}
		\label{fig:complx2000_num}       
	\end{figure*}

\begin{figure}[tb]
		\centering
		\includegraphics[clip, trim=4.5cm 10.5cm 5cm 11cm,width=0.63\textwidth]{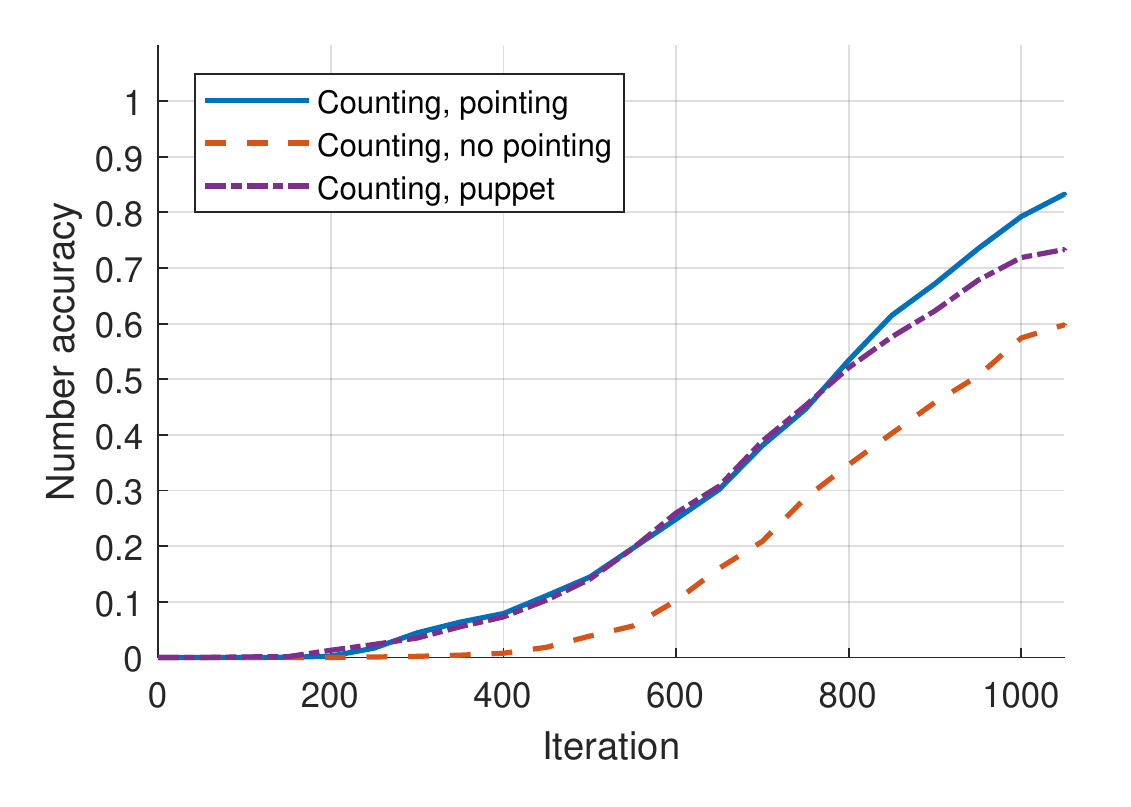}
		\caption{Accuracy of counting development (number output accuracy), average from 30 simulations. The model was trained to count in described in this section manner assuming training length of 1050 iterations. Tested in 3 conditions: when pointing required, when pointing forbidden and when counting with a pointing puppet.}
		\label{fig:complx1050_num_1}       
	\end{figure}

	Fig.~\ref{fig:probability} shows the probability of each simulated skill being trained during the training session. As visible in the figure, there are no values included at the horizontal axis. We did not know how long the training should take. We assumed that we should finish the training when the results will be at a similar level to those from 4-year-old children. The values for children for counting with pointing, without it, and with a pointing puppet are 82.5\%, 50\% and 77.5\%\footnote{Those values represent the situation when children were counting objects from a smaller set: 7-10 (as it is closer to the one we use). We considered only pointing conditions and not the touch one which is analysed later. The values were expressed as the percentage of correct answers} \citep{alibali1999function}. The average of that is 70\%. Thus, we tried to adjust the length of our training so that the average of our results is also 70\%. First, we run the training through 2 epochs. As visible in the Fig.~\ref{fig:complx2000_num}, the final result is too high (the average value is equal to 93\%). We can not just use the results from the middle of that training session as the probability distribution function (how often each skill is used in training) was assumed to finish together with the training. Thus, we tried several shorter training sessions and we found 1050 iteration training to have the average result closest to 70\% (it was 72.1\%). As you can see in Fig.~\ref{fig:complx1050_num_1}, the character of the plot is different than the one for the first 1050 iterations from the previous figure.
	
\begin{figure}[tb]
		\centering
		\includegraphics[clip, trim=4.5cm 10.5cm 5cm 11cm,width=0.63\textwidth]{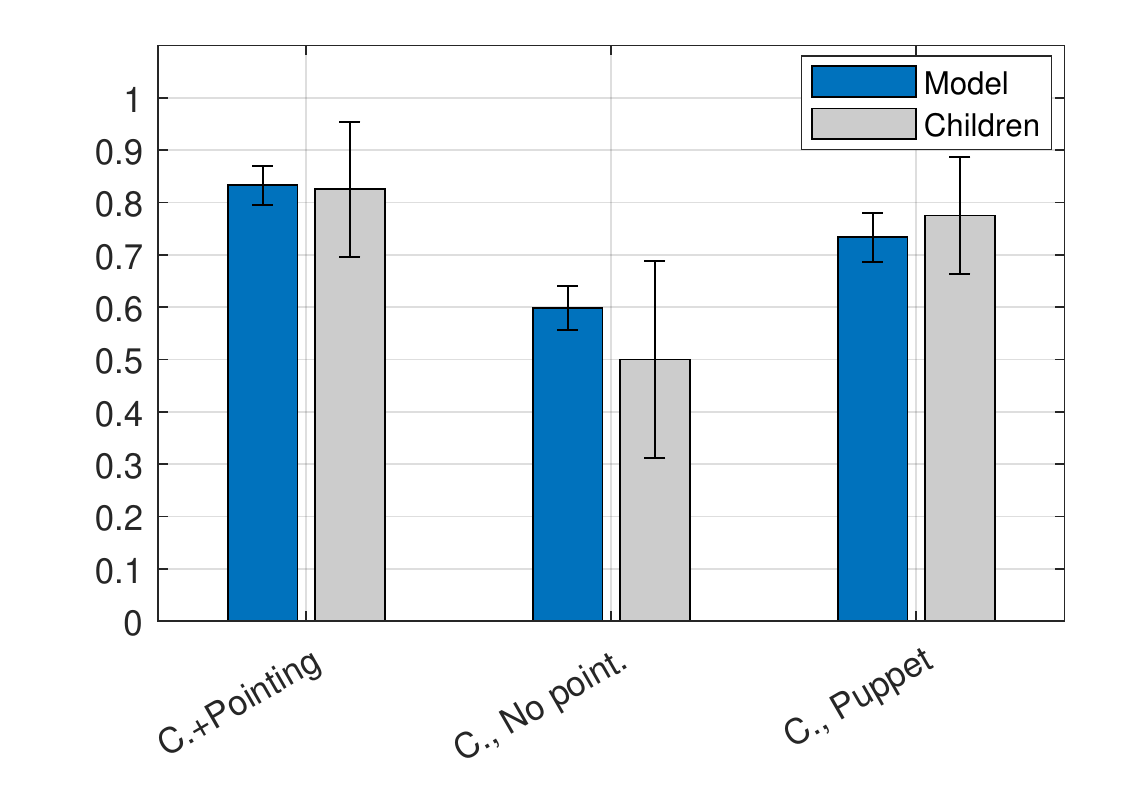}
		\caption{Accuracy of counting (number output accuracy) from 30 simulations after 1050 iterations compared with the results for children from \cite{alibali1999function}. There were 3 conditions considered: when pointing required, when pointing forbidden and when counting with a pointing puppet. The error bars indicate 95\% confidence intervals.}
		\label{fig:complx1050_num_2}       
	\end{figure}

	We compared the final results for this simulation with those from \cite{alibali1999function}. There was a big similarity between our results and those from children studies (see Fig.~\ref{fig:complx1050_num_2}). We also obtained the highest result of the counting performance for counting with pointing and the worst for counting without it. The results for counting with a puppet are relatively similar to the results with pointing (on average a bit worse). This was true for these and human studies.
	
	A one-way ANOVA analysis showed a strong statistical difference between the results from our 3 conditions, $F(2,87)=30.13$, $p < 0.001$. We conducted a Post Hoc (Tukey’s honestly significant difference) test pairwise. We found that the difference between counting with pointing and without it as well as between counting with a puppet and without pointing is strongly statistically significant ($p < 0.001$). There was also a statistical difference ($p = 0.004$) between the results of counting with pointing and counting with a puppet (in the case of studies with children, that difference was not statistically significant).
	
\begin{figure}[tb]
		\centering
		\includegraphics[clip, trim=4.5cm 10.5cm 5cm 11cm,width=0.63\textwidth]{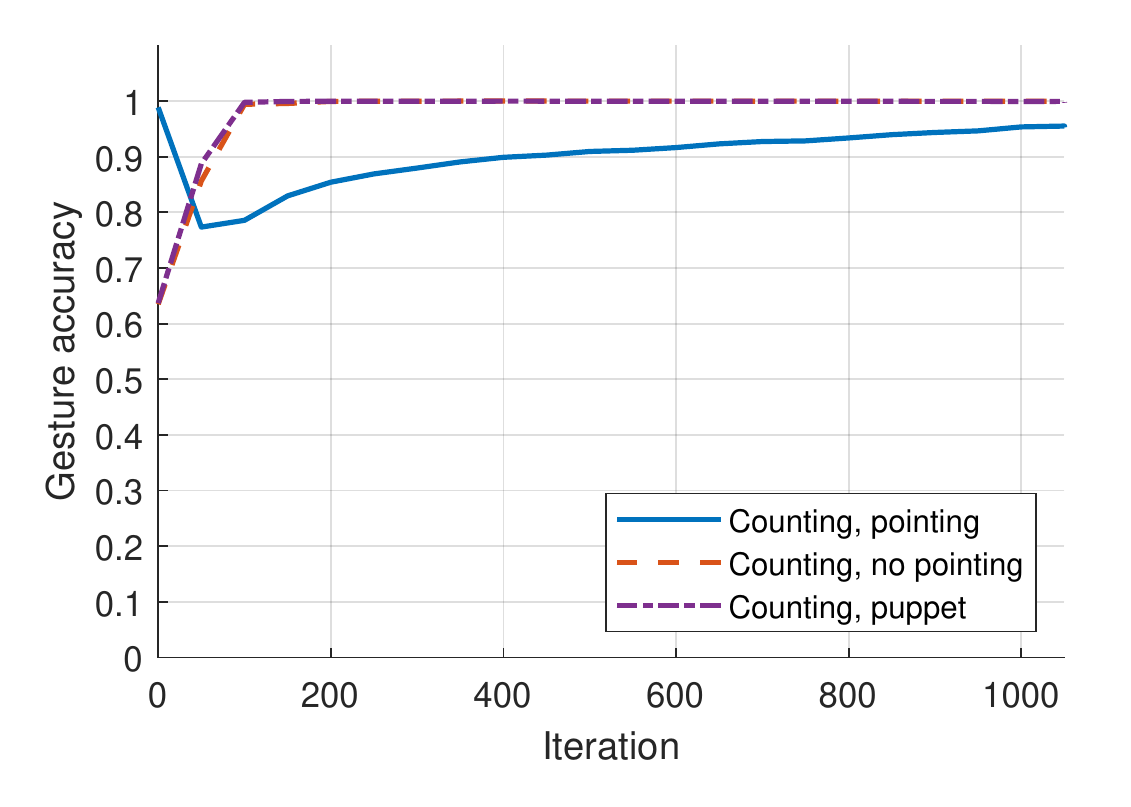}
		\caption{Accuracy of pointing development (gesture output accuracy), average from 30 simulations. he model was trained to count in described in this section manner assuming training length of 1050 iterations. Tested in 3 conditions: when pointing required, when pointing forbidden and when counting with a pointing puppet.}
		\label{fig:complx1000_gest}       
	\end{figure}

	The gesture performance, we can see in Fig.~\ref{fig:complx1000_gest}. Similarly, as in Study 2 the network almost immediately learns to stay in the base position (in the case of 5 and 6 skill).
	
	To compare the results with those from \cite{alibali1999function} we performed t-tests between corresponding observations (comparing model results with results from children). We found that for all of the conditions the results are not significantly different than the results obtained with children:
	\begin{itemize}
		\item Counting with pointing: $t(48)=0.13$, $p = 0.896$
		\item Counting without pointing: $t(48)=1.19$, $p = 0.240$
		\item Counting with a puppet: $t(48)=0.76$, $p = 0.450$
	\end{itemize}

	We can conclude that our model performance for particular simulated skills is similar to the corresponding counting performances of 4-year-old children.
	
	\subsection{Distance to counted objects}\label{dist}
	
	It has been found that the performance of counting is higher when children can touch the objects (even higher than in the case of pointing). A similar finding was observed in the case of puppet touch \citep{alibali1999function}. They came with an assumption: ``these findings suggest that the key difference between touching and pointing is not the tactile information provided by the touch, but rather the distance between the indication act and the object indicated'' \citep[][p. 47]{alibali1999function}.
	
	We performed some experiments to see if we will be able to observe differences in the results when the network points to closer objects and if the counting performance is improved by that. To do so, we tested the model trained in the same way as described in Subsection \ref{count_complex} in two cases:
	\begin{itemize}
		\item objects are in two lowest rows - closer to the hand
		\item objects are in two highest rows - farther from the hand
	\end{itemize}
	This distinction was done only in the test sets, the training set constituted of objects distributed throughout all 5 rows.
		
\begin{figure*}[tb]
		\centering
		\includegraphics[clip, trim=0.5cm 10.5cm 0.5cm 11cm,width=1\textwidth]{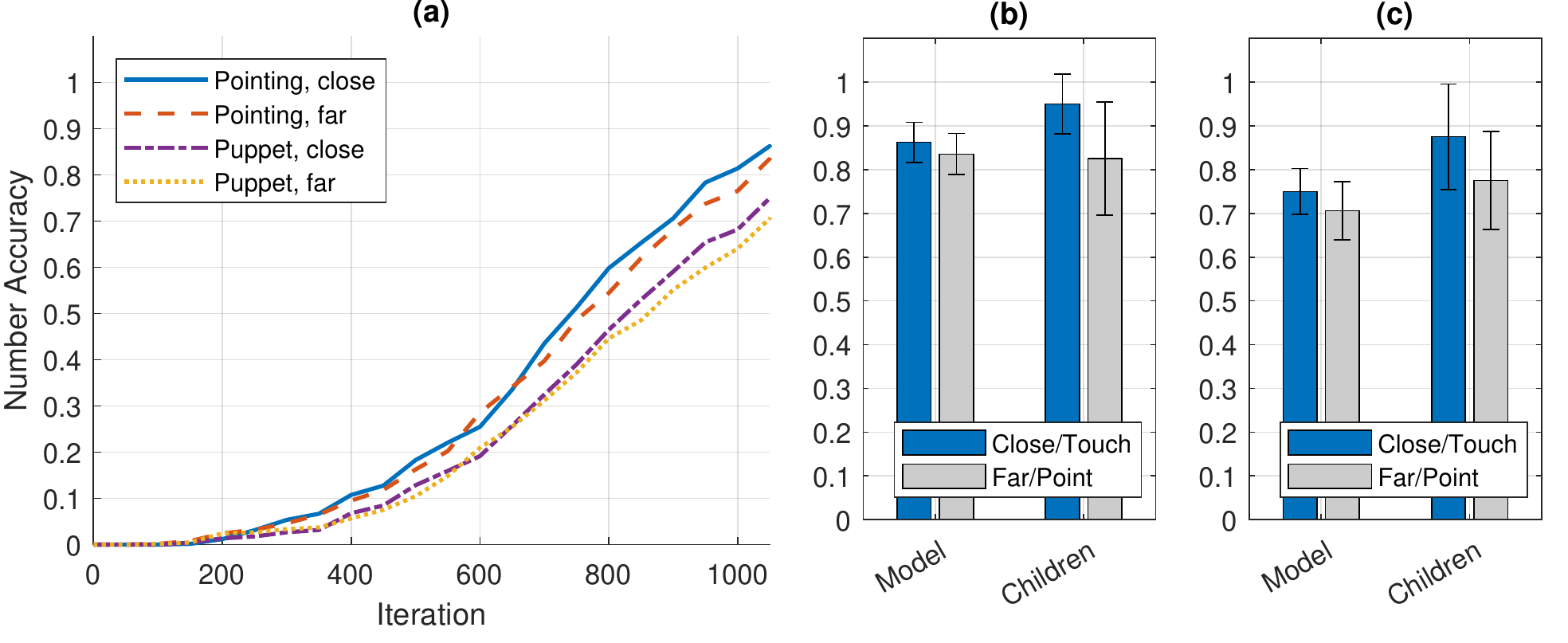}
		\caption{Accuracy of counting (number output accuracy), average from 30 simulations. The model trained as described in Subsection \ref{count_complex}. Tested in 3 conditions: when pointing required, when pointing forbidden and when counting with a pointing puppet. The error bars (in the bar plots) indicate 95\% confidence intervals. (a) Development curves (test set used every 50 iterations). (b) Final accuracy (after 1050 iterations) for counting with pointing. (c) Final accuracy for counting with a pointing puppet.}
		\label{fig:complx2000_far_close}       
	\end{figure*}

	As can be seen in Fig.~\ref{fig:complx2000_far_close}, the results are better when the objects are closer to the hand (lower rows). When we run t-test for the model results we can see that for both simulated skills: pointing and a pointing puppet, those differences are not statistically significant: $t(58)=0.81$, $p = 0.420$ and $t(58)=1.02$, $p = 0.311$ respectively. Comparing to the study of \cite{alibali1999function} with children we can find that results where children touch the objects are better compared to those where they only point but the difference after running t-test for those result (for smaller sets - under 10) is also not statistically significant\footnote{We compared our results with those of small sets, the authors in \cite{alibali1999function}, however, found a statistical difference as they considered both ranges (small and large sets) together (having larger number of samples). They also used a different test repeated measures ANOVA.} ($p > 0.05$): $t(38)=1.96$, $p = 0.056$. The same we can see in the case of the pointing puppet condition: $t(38)=1.30$, $p = 0.201$.
	
	We can conclude that both the model and the children tend to better count closer items (or touched ones in the case of children) but the influence of that condition change is not significant.

	\subsection{Size of a counted set}
	In our experiment the counted sets were of size from 1 to 10, i.e. in the input image the number of tennis balls was between 1 and 10. The experiment on children conducted by \cite{alibali1999function} was done for different sizes of counted sets (from 7 till 17). In the case of our work, we were limited due to technical reasons (range of camera view and iCub joint angles). Increasing the amount of objects would require some significant changes in the experiment, like the rotation of the robot's head or eyes, thus, no static image (and much more complex task).
	
	Because of these limitations, the comparison between different set sizes is not reflecting the child experiment i.e. small and big sets are covering a different number of objects in our experiment compared to the one with children.
	
\begin{figure}[tb]
		\centering
		\includegraphics[clip, trim=4.5cm 8cm 5cm 8cm,width=0.63\textwidth]{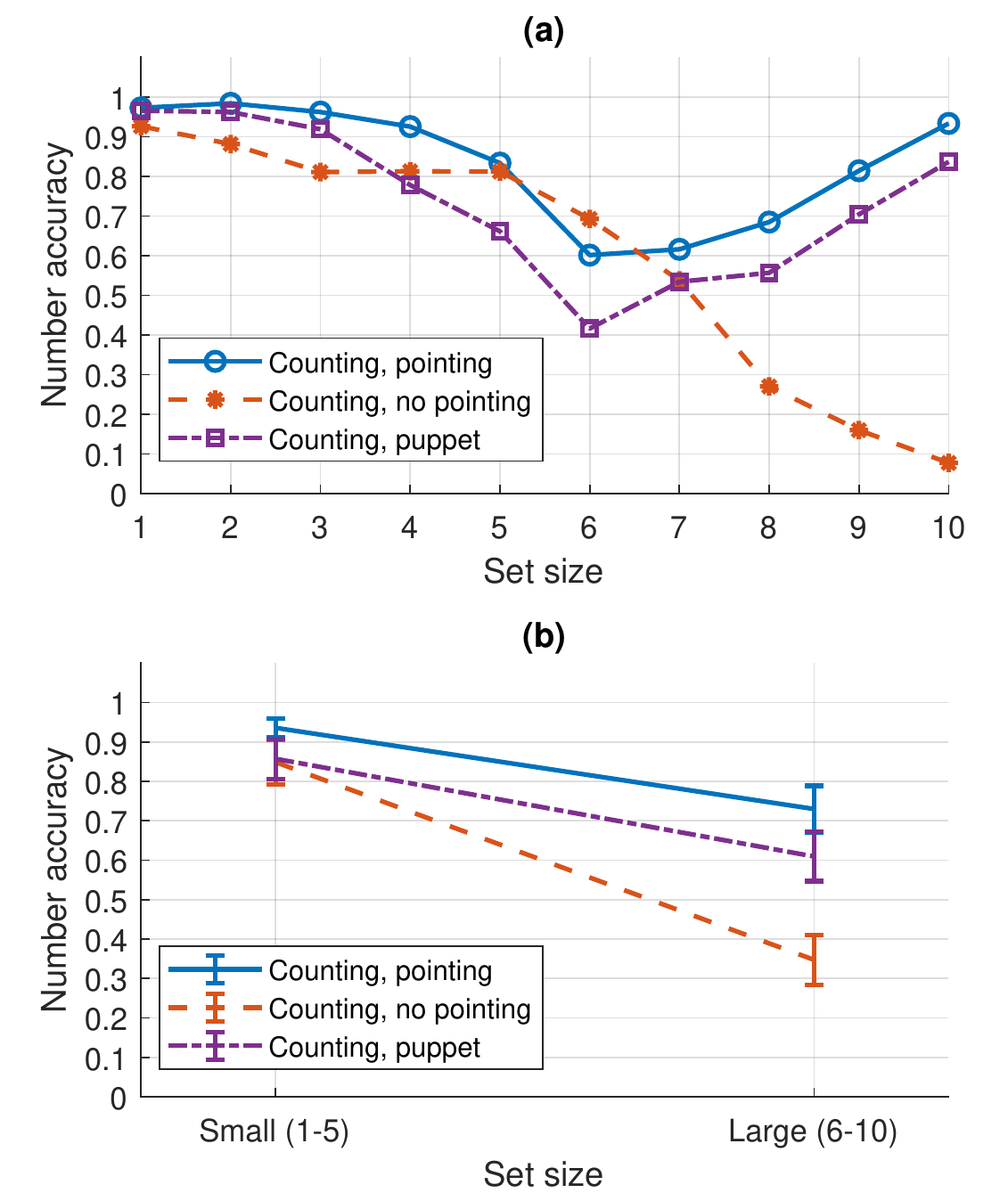}
		\caption{Accuracy of counting for sets of a different size, average from 30 simulations. The model was trained as described in Subsection \ref{count_complex}. Tested in 3 conditions: when pointing required, when pointing forbidden and when counting with a pointing puppet. (a) The values for each numerosity. (b) The values for small and large sets. The error bars indicate 95\% confidence intervals.}
		\label{fig:Set_size}       
	\end{figure}

	A general observation in the study with children was that the smaller sets are counted better. In Fig.~\ref{fig:Set_size} we present the final (after the training session was finished) accuracy of counting (of the model) for a specific number of objects as well the summary results when we grouped them in small numerosity sets (1 to 5) and the big ones (6 to 10)\footnote{In the case of the experiment with children the small sets were covering 7 to 10 objects and big ones 13 to 17}. Pairwise t-tests showed that for each of the simulated skill the results are better for smaller sets ($p < 0.001$): counting with pointing: $t(58)=6.39$; counting without pointing: $t(58)=11.55$; counting with a puppet: $t(58)=6.02$.
		
	The model results are in line with the results from studies with children where a better performance was observed in the case of smaller sets for the corresponding counting tasks as well \citep{alibali1999function}.
	
\section{Summary and conclusions}

This paper presents a developmental neuro-robotics model capable of producing pointing gestures while counting real items. The model uses pointing which help to keep track of counted objects. A deep neural network, composed of a Convolutional part and a Recurrent part, is used to generate number words and joint angle values (pointing). A robotic platform is used as an embodiment providing the visual information and sensorimotor data for the training.

The experimental results of the realistic testing of the robotic model were compared with studies conducted with children, particularly with \cite{alibali1999function}. We observed that our model resembled many similarities from the point of view of counting performance with 4-year-old children. We can see that the performance of our model when it is trained in a particular way replicated the results of children to the point where they are statistically not different from them (however, the counted set sizes were different). It was observed that the best performance is obtained when the model is pointing to the items itself and the worse when it was not allowed to point. The condition with a pointing puppet did not improve the counting (compared to self pointing), even though, pointing, in that case, was conducted perfectly. Moreover, such a condition was slightly worse compared to the condition where the model was pointing by itself. Those observations are fully in line with those from studies on children \citep{alibali1999function}. We found that the model tends to have better results when the counted items are located closer to the hand (in the case of model pointing by itself and a pointing puppet). Assuming that the main difference between touching and pointing in the case of children comes from the distance between a hand and the item \cite[as suggested in][]{alibali1999function}. We can, again, see similar results in children. We also observed the trend of counting smaller sets better than bigger ones. Which was also true for children.

What was an interesting observation is that our results vary depending on what training procedure the model followed. The results where we trained our network to count only with pointing, as well as where it was trained equally to count with and without pointing and with a pointing puppet, were significantly different than those from children studies. This might suggest that also in the case of children the training methodology is very important and can determine in which of these conditions they will count better.

Based on the obtained results, an important question that can be posed is: does the model really count? The answer to that question is not as straight forward is it may seem. Researchers defined several counting tests to examine children's ability to use counting principles and their number sense \citep{le2007one,wynn1992children}, two most common are: ``How many?'' and ``Give a number'' tests. In the first one, children were supposed to count the objects and give the answer of how many items were presented to them. The second one, required them not only to count but also to provide the examiner with a specific number of objects.
We can conclude that the presented model can pass the ``How many?'' test up to the numerosity of 10 (if trained long enough) and in the meaning of that test, it can count. However, the question if the model is able to perceive the numerosity and to what level, is more difficult to answer. To understand to what level proposed model understands quantity, we would have to conduct a more detailed analysis of the internal activation of the network and this would still not give as a straight forward answer if this is a real understanding. However, as visible in studies of \cite{le2006re,le2007one,wynn1990children,wynn1992children} the test for a real understanding of numerosities was not easy to define in children as well.

In future work, it would be interesting to analyze the internal activation of the network which could answer some additional questions, such as why the network performance was the worst in a pointing puppet condition in Study 1. Also, there could be some other more human-like representation of a number output used (phonetic based) and an internal representation of numbers based on a logarithmic scale like in \cite{dehaene1990numerical,dehaene1992varieties} or a linear one with a scalar variability \citep{gallistel1992preverbal,gallistel2000non} which, as shown in comparison tasks, seem to be more human-like representations. Other more recent representations that could be used which are taking into account some effects observed in humans are presented in \cite{rotondaro2019number, gigliotta2019midpoint}.



\section{Compliance with Ethical Standards}
\paragraph{Acknowledgement}
This is a pre-print of an article published in Psychological Research. The final authenticated version is available online at: \url{https://doi.org/10.1007/s00426-020-01428-8}

\paragraph{Funding}
The authors acknowledge the UK EPSRC support through the project grant EP/P030033/1 (NUMBERS) and the EU support via the H2020 Marie Sk\l{}odowska-Curie ITN APRIL project (grant agreement No 674868).
\paragraph{Conflicts of interest}
The authors declare that they have no conflict of interest.

\paragraph{Ethical approval}
This article does not contain any studies with human participants or animals performed by any of the authors.
\paragraph{Availability of data and material}
The datasets generated during the current study are available from the corresponding author on reasonable request.
\paragraph{Code availability}
The code created in the current study is available from the corresponding author on reasonable request.

%
%
\bibliographystyle{apalike} 
\bibliography{ref,ref_adn}   

\end{document}